\documentclass{article}
\usepackage{spconf,amsmath,epsfig}

\let\OLDthebibliography\thebibliography
\renewcommand\thebibliography[1]{
  \OLDthebibliography{#1}
  \setlength{\parskip}{0pt}
  \setlength{\itemsep}{0pt plus 0.3ex}
}

\usepackage{amsmath}
\usepackage{amssymb}
\usepackage{gensymb}
\usepackage{booktabs}
\usepackage{array}

\usepackage{tabularx}
\newcolumntype{L}[1]{>{\raggedright\arraybackslash}m{#1}}
\newcolumntype{C}[1]{>{\centering\arraybackslash}p{#1}}

\begin{document}\sloppy

\def\x{{\mathbf x}}
\def\L{{\cal L}}

\title{Disjoint Contrastive Regression Learning for Multi-Sourced Annotations}
%
\name{Xiaoqian Ruan$^1$, Gaoang Wang$^{1}$}
\address{$^1$ Zhejiang University-University of Illinois Urbana-Champaign Institute, Zhejiang University, China\\
\{xiaoqianruan, gaoangwang\}@intl.zju.edu.cn}

\maketitle

\begin{abstract}
Large-scale datasets are important for the development of deep learning models. Such datasets usually require a heavy workload of annotations, which are extremely time-consuming and expensive. 
To accelerate the annotation procedure, multiple annotators may be employed to label different subsets of the data. However, the inconsistency and bias among different annotators are harmful to the model training, especially for qualitative and subjective tasks. 
To address this challenge, in this paper, we propose a novel contrastive regression framework to address the disjoint annotations problem, where each sample is labeled by only one annotator and multiple annotators work on disjoint subsets of the data. To take account of both the intra-annotator consistency and inter-annotator inconsistency, two strategies are employed. Firstly, a contrastive-based loss is applied to learn the relative ranking among different samples of the same annotator, with the assumption that the ranking of samples from the same annotator is unanimous. Secondly, we apply the gradient reversal layer to learn robust representations that are invariant to different annotators.
Experiments on the facial expression prediction task, as well as the image quality assessment task, verify the effectiveness of our proposed framework. 
\end{abstract}
\section{Introduction}
\label{sec:intro}

Large-scale labeled datasets are in heavy demand to boost the performance of deep learning models, such as object retrieval \cite{chen2014ranking}, tracking \cite{li2011graph}, crowd counting \cite{jiang2019learning}, and multi-modal learning \cite{summaira2021recent}. Such datasets often require a high cost of annotations, which are extremely time-consuming and expensive.
Moreover, a large-scale dataset is usually labeled by multiple annotators. Due to individual preferences, the labeled results are not always consistent among different annotators, especially for qualitative and subjective tasks. As a result, to reduce the annotator bias, each sample is often required to be labeled by multiple annotators, followed by an average or a majority vote to obtain the final labels, which leads to a heavier cost of annotation. 
To alleviate the labeling effort, it is more efficient to assign disjoint subsets of the data to individual annotators for labeling. Therefore, how to address the challenge resulting from the inconsistency and bias among different annotators becomes a tough task.

\begin{figure}[!t]
\begin{center}
\includegraphics[width=0.9\linewidth]{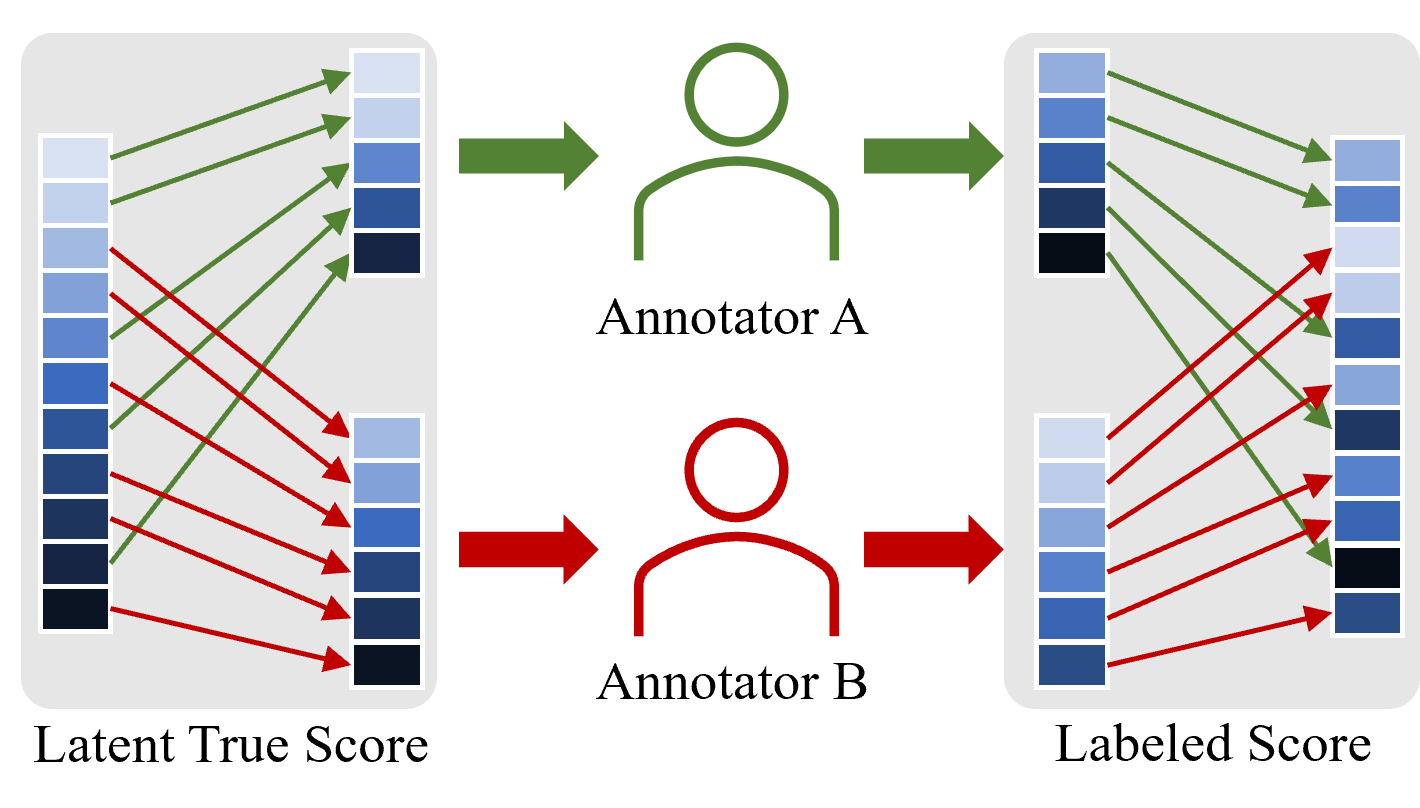}
\end{center}
   \caption{Example of the intra-annotator consistency and inter-annotator inconsistency phenomenon. Each box represents a sample from the dataset. The blue color from dark to light represents the score from low to high. After labeling, the intra-annotator consistency remains, while the inter-annotator consistency is violated.}
\label{fig:inconsistency}
\end{figure}

Some approaches \cite{rodrigues2018deep,xu2019learning,wang2020representation,tanno2019learning,keswani2021towards,guan2018said} deal with the noisy-label issue or labeling biases from crowd-sourced annotations. Many advances \cite{liu2020early,yao2019safeguarded,bahri2020deep,algan2021image} have been made in recent years, especially for classification tasks with crowd-sourced labels. For regression tasks, except for the inconsistency among multiple annotators, the intra-annotator consistency also needs to be considered, in which the labeled rankings of samples from the same annotator usually align well with the rankings of the latent ground truth scores. An example of this phenomenon is shown in Fig.~\ref{fig:inconsistency}. How to take account of both intra-annotator consistency and inter-annotator inconsistency for regression tasks is still underexplored.

\begin{figure*}[!t]
\begin{center}
\includegraphics[width=1.0\linewidth]{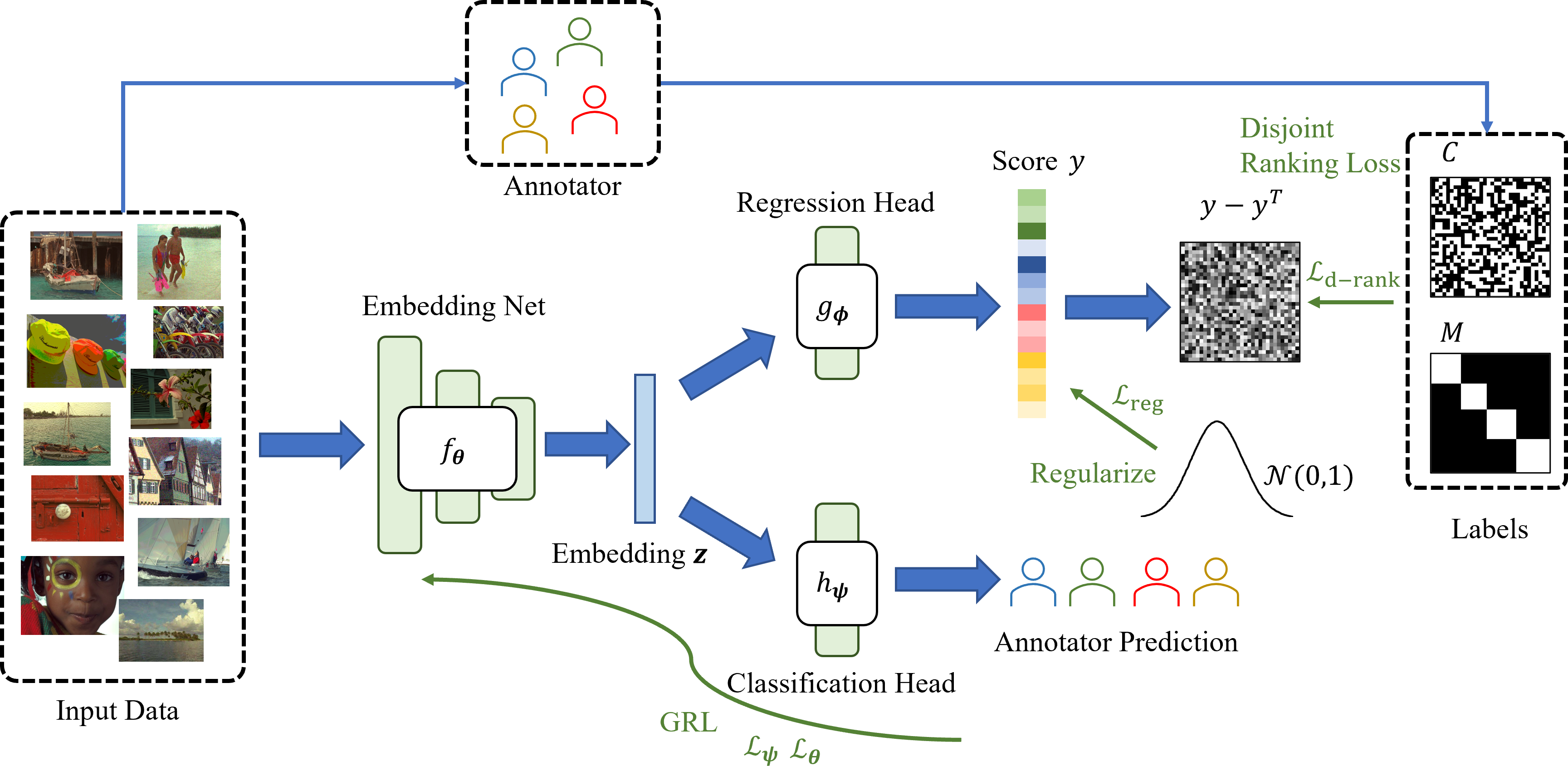}
\end{center}
   \caption{The framework of disjoint contrastive regression (DCR). The input data is first fed to the embedding net $f_{\boldsymbol{\theta}}$. The score is then predicted by the regression head $g_{\boldsymbol{\phi}}$. The pairwise distance of the prediction is optimized by the supervision of disjoint ranking loss. In the meanwhile, the embeddings are fed to a second classification branch $h_{\boldsymbol{\psi}}$ that aims to classify different annotators. We adopt the gradient reversal layer (GRL) to learn the robust annotator-invariant embeddings.}
\label{fig:flowchart}
\end{figure*}

In this paper, we present a novel approach for regression tasks from disjoint annotations with the designed disjoint margin ranking loss and gradient reversal layer (GRL) \cite{ganin2015unsupervised}. Specifically, We extend the margin ranking loss based on the facts of disjoint annotations, with both the intra-annotator consistency and inter-annotator inconsistency adopted, leading to a robust embedding framework. Besides that, regularization of the batch distribution is employed to avoid the collapse to naive solutions. To learn annotator invariant representations, GRL is further adopted in the model training. All these make up the entire framework of our proposed Disjoint Contrastive Regression (DCR). A demonstration of the framework is shown in Fig.~\ref{fig:flowchart}.
The main contributions of this paper are summarized as follows: 
\begin{itemize}
\setlength{\itemsep}{0pt}
\setlength{\parsep}{0pt}
\setlength{\parskip}{0pt}
\item We present a novel contrastive learning framework for regression tasks with disjoint annotations.
\item We extend the margin ranking loss and resolve the intra-annotator consistency and inter-annotator inconsistency.
\item We adopt GRL in the model training for learning annotator invariant embeddings. 
\item We conduct experiments on the facial expression prediction task and image quality assessment task, demonstrating the effectiveness of the proposed DCR.
\end{itemize}

\section{Related Work}

\noindent \textbf{Crowd-Sourced Annotations.}
The inter-and intra-observer variations among experts due to intrinsic differences in human annotators widely occur in computer vision tasks. Many approaches \cite{rodrigues2018deep,xu2019learning,wang2020representation,tanno2019learning,keswani2021towards,guan2018said,wang2021dail} have been proposed to deal with the inconsistent label problems from the crowd-sourced annotations. 
For example, \cite{chao2018cross} proposes an EM algorithm and a crowd layer to deal with the crowd-sourcing problems. 
To solve the noise and inconsistency in some limited-sized datasets from multiple annotators, \cite{xu2019learning} proposes a crowd-sourced representation learning framework that generates a large number of training instances from only small-sized labeled sets. 
However, most existing works focus on classification tasks. How to deal with inconsistent annotations with numerical labels for regression tasks is still underexplored.

\noindent \textbf{Cross-Dataset Adaptation.}
Since data from source and target domains have different distributions, the key challenge for domain adaptation \cite{pei2018multi,cao2018partial,kang2019contrastive,zhao2019learning,wang2018deep,shao2020domain,yang2020fda} is to address the domain shift between them. Fine-tuning \cite{ge2017borrowing,guo2019spottune}, which diminishes the shift between the two domains, is a typical way for domain adaptation with transfer learning \cite{kouw2018introduction,liu2016transferring}.
For instance, \cite{guo2019spottune} proposes an adaptive fine-tuning approach to find the optimal strategy per image for the target data. Transfer learning with triplet loss \cite{liu2016transferring} is employed for bridging the gap between different domains.
Adversarial learning is another essential way for domain adaptation \cite{pei2018multi,cao2018partial,zhang2019domain,vu2019advent,chen2020adversarial}. 
For example, \cite{cao2018partial} presents a partial adversarial domain adaptation approach to alleviate negative transfer by reducing the weights of outliers and promote positive transfer. 
A multi-adversarial domain adaptation approach \cite{pei2018multi} is proposed to capture multi-mode structures to ensure fine-grained alignment of distributions. The domain-symmetric network \cite{zhang2019domain} is presented on a symmetric design of source and target classifiers with adversarial training. However, such domain adaptation approaches cannot directly address the inconsistency annotation problem raised from multi-sourced data.

\section{Method}
\subsection{Problem Description}
Assume the whole dataset $\mathcal{X}$ are labeled by $K$ annotators. Denote the subset annotated by the $k$-th annotator as $\mathcal{X}_k$ and different subsets are disjoint with each other, \textit{i.e.}, $\mathcal{X}_k \cap \mathcal{X}_j = \O$ if $j \neq k$ and $\mathcal{X} = \cup_{k=1}^{K} \mathcal{X}_k$.
Given the set of labeled samples, we aim to learn a robust embedding network $f_{\boldsymbol{\theta}}: \boldsymbol{x} \rightarrow \boldsymbol{z}$, as well as a regression head $g_{\boldsymbol{\phi}}: \boldsymbol{z} \rightarrow y$, based on the supervision of the disjoint and inconsistent annotation $\Tilde{y}$, where $\boldsymbol{x}$, $\boldsymbol{z}$ and $y$ are the input sample, representation, and label, respectively.

\subsection{Preliminary on Margin Ranking Loss}
Margin ranking loss is commonly used for contrastive regression problems, which is defined as follows,
\begin{equation}
\begin{aligned}
& \mathcal{L}_{\text{rank}} = \frac{1}{N}\sum_{i,j} \left[\gamma-\boldsymbol{C}_{i,j}({y}_i-{y}_j)\right]_+,
\label{eq:rank_reg}
\end{aligned}
\end{equation}
where $N$ is the batch size; $y$ is the prediction; $\boldsymbol{C}_{i,j}=1$ if the annotated scores satisfy $\Tilde{y}_i \geq \Tilde{y}_j$, otherwise $\boldsymbol{C}_{i,j}=-1$; $\gamma$ is a pre-defined margin, and $[\cdot]_+$ clamps values to be non-negative. By minimizing the margin ranking loss in the training, the predicted scores should follow the pairwise relative rankings of the labeled scores.

\subsection{Disjoint Contrastive Regression (DCR)}
There are three drawbacks of the commonly used margin ranking loss. First, it omits the issue caused by inter-annotator inconsistency. It is harmful to optimize the relative ranking when two samples are labeled by different annotators. Second, it can easily collapse to naive solutions by only calculating the relative order with no absolute scores as reference. Third, it does not consider the domain shift coming from different subsets. 
To deal with the issues above, we firstly extend the margin ranking loss to the proposed disjoint ranking loss with conditional annotator indicators, to avoid the direct comparison across annotators. Then, regularization on the batch predictions is employed to ensure relative ranking following a certainly designed distribution. Thirdly, we reduce the preference bias across annotators with the gradient reversal layer (GRL) with adversarial training, leading to more bias-invariant representations. The proposed disjoint contrastive regression (DCR) framework is shown in Fig.~\ref{fig:flowchart}. The details are demonstrated as follows. 

\noindent \textbf{1) Disjoint Margin Ranking Loss.}
Given a sample $\boldsymbol{x}_i$ annotated by the $k$-th annotator, denote the annotated label as $\Tilde{y}_i^k$, and the latent unbiased true label as $y_i^*$. Usually, $y_i^*$ is not accessible. Considering regression tasks, due to the preference bias of different annotators, it is quite common that, for two samples annotated by different annotators, we have $\Tilde{y}_i^k < \Tilde{y}_j^l$ while the latent true label is $y_i^{*} > y_j^{*}$ in some situations. Such inconsistency among disjoint annotations can be harmful to the model training.
Meanwhile, we assume the ranking of labels within the same annotator usually keeps consistent. Based on the above two phenomenons, we should alleviate the inconsistency for inter-annotator labels, and utilize the consistency of intra-annotator labels for representation learning. 

To this end, we extend the margin ranking loss Eq.~(\ref{eq:rank_reg}) to the proposed disjoint margin ranking loss as follows,
\begin{equation}
\begin{aligned}
& \mathcal{L}_{\text{d-rank}} = \frac{1}{N}\sum_{i,j} \boldsymbol{M}_{i,j} \left[\gamma-\boldsymbol{C}_{i,j}({y}_i-{y}_j)\right]_+,
\label{eq:d-rank}
\end{aligned}
\end{equation}
and $\boldsymbol{M}_{i,j}$ is defined as,
\begin{equation}
\begin{aligned}
& \boldsymbol{M}_{i,j} = \sum_k \boldsymbol{1}_{[i \in A_k]} \boldsymbol{1}_{[j \in A_k]},
\label{eq:m}
\end{aligned}
\end{equation}
where $\boldsymbol{1}_{[i \in A_k]}$ is the indicator function and returns 1 if the sample $i$ is from the $k$-th annotator $A_k$, otherwise it returns 0. $\boldsymbol{M}_{i,j}$ is the judgment of whether the sample $i$ and the sample $j$ are labeled by the same annotator.
In other words, $\boldsymbol{M}_{i,j}$ ensures that the loss is optimized within the same annotator. As a result, Eq.~(\ref{eq:d-rank}) will generate a penalty if the predicted pairwise relative ranking is conflicted with the annotated pairwise relative ranking. With Eq.~(\ref{eq:d-rank}), we can avoid cross-annotator inconsistency and utilize the intra-annotator consistency at the same time. 

\noindent \textbf{2) Regularize on the Relative Ranking.}
Eq.~(\ref{eq:d-rank}) only compares the relative order given two samples, which is unstable and easy to collapse to extreme naive solutions. 
To address such a problem, we constrain the predictions to follow the normal distributions $\mathcal{N}(0, 1)$ within batch samples, \textit{i.e.},
\begin{equation}
\mathcal{L}_{\text{reg}} = \left(\frac{1}{N}\sum_i {y}_i\right)^2
+\left(\frac{\sum_i {y}_i^2}{N-1}-1\right)^2,
\label{eq:reg}
\end{equation}
where the first term constrains the distribution to be zero-mean, while the second term regularizes the variance to be close to 1. With the normal distribution as a reference, the regularization term stabilizes the training, which is verified in the experiment section.

\noindent \textbf{3) Invariant Learning with GRL.}
To further improve the robustness of regression performance across multiple biased disjoint annotations, bias-invariant learning is crucial to ensure that the latent representations are not annotator-dependent. As a result, we adopt one commonly used domain adaptation strategy with the gradient reversal layer (GRL) \cite{ganin2014unsupervised} for learning the latent embeddings.
In addition to the embedding network $f_{\boldsymbol{\theta}}: \boldsymbol{x} \rightarrow \boldsymbol{z}$ and the regression head $g_{\boldsymbol{\phi}}: \boldsymbol{z} \rightarrow y$, we consider another annotator classifier $h_{\boldsymbol{\psi}}: \boldsymbol{z} \rightarrow \boldsymbol{p}$, where $\boldsymbol{p}$ is the predicted probability for classifying annotators after the soft-max operation.
We consider two loss functions, 
\begin{equation}
\mathcal{L}_{\boldsymbol{\psi}} = -\frac{1}{N}\sum_i \boldsymbol{p}_i^*\log h_{\boldsymbol{\psi}} \left(f_{\hat{\boldsymbol{\theta}}}(\boldsymbol{x}_i) \right),
\label{eq:grl1}
\end{equation}
\begin{equation}
\mathcal{L}_{\boldsymbol{\theta}} = \frac{1}{N}\sum_i \boldsymbol{p}_i^*\log h_{\hat{\boldsymbol{\psi}}} \left(f_{\boldsymbol{\theta}}(\boldsymbol{x}_i) \right),
\label{eq:grl2}
\end{equation}
where $\boldsymbol{p}^* \in \mathbb{R}^K$ is the one-hot vector as the annotator indicator; $\hat{\boldsymbol{\theta}}$ and $\hat{\boldsymbol{\psi}}$ 
denote the frozen version of the parameters. In Eq. (\ref{eq:grl1}), we freeze the embedding network and learn the classification head to minimize the cross-entropy, while in Eq. (\ref{eq:grl2}), we freeze the classification head and optimize the embedding network. Such adversarial training strategy learns bias-invariant representations.

\noindent \textbf{4) Combined Loss for  Representation Learning.} 
We summarize the total loss of the proposed DCR as follows,
\begin{equation}
\mathcal{L} = \mathcal{L}_{\text{d-rank}}+\lambda_1 \mathcal{L}_{\text{reg}}+\lambda_2 \mathcal{L}_{\boldsymbol{\psi}}+\lambda_3 \mathcal{L}_{\boldsymbol{\theta}},
\label{eq:total_loss}
\end{equation}
where $\lambda_1 \sim \lambda_3$ are the weights of individual losses, all set to 0.2 as default.

\section{Experiments}

\subsection{Implementation Details}


\subsubsection{Datasets}
\begin{figure}[t]
    \centering
    \includegraphics[width=1.0\linewidth]{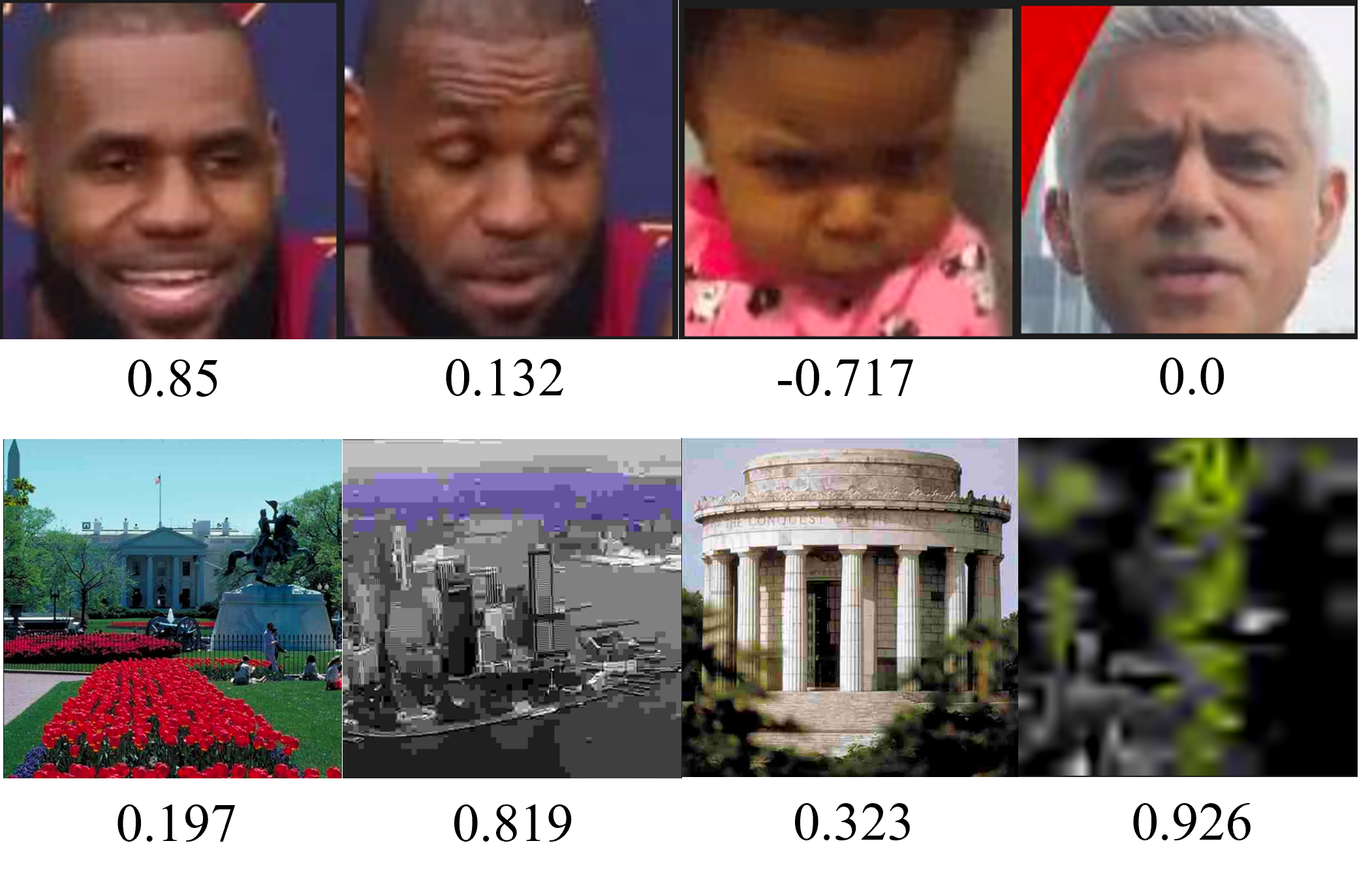}
    \caption{Selected examples with their corresponding scores, including the images and their corresponding scores, from the datasets. The images on the first line are samples for facial expression score prediction while those on the second line are for IQA.}
    \label{fig:galaxy}
\end{figure}

We evaluate the performance of our method on two regression tasks, namely facial expression score prediction task with AFEW-AV \cite{kossaifi2017afew} and Aff-Wild \cite{zafeiriou2017aff} datasets, and image quality assessment (IQA) task with CSIQ \cite{larson2010most} and TID2013 \cite{ponomarenko2013color,ponomarenko2013new} datasets. For all these four datasets, 80\% samples are used for training, while the remaining samples are used for testing. Some examples from the datasets are shown in Fig.~\ref{fig:galaxy}. We present the details of these four datasets as follows.

\textbf{AFEW-AV} is a subset of the Acted Facial Expressions in the Wild (AFEW) \cite{dhall2012collecting} dataset, focusing on 7 basic facial expressions, namely disgust, fear, anger, sadness, happiness, surprised and neural. Apart from these classification labels, AFEW-AV provides the discrete valence and arousal scores, ranging from -10 to 10. This dataset contains 600 ``in-the-wild'' video clips with variations of backgrounds and poses of the subjects. For simplicity, we use the individual frames as distinct samples for the regression task.

\textbf{Aff-Wild} dataset consists of around 300 videos with 200 subjects, including 130 males and 70 females. This dataset also provides two continuous emotion dimensions, in terms of valence and arousal, which range from -1 to 1. 

\textbf{CSIQ} dataset is based on 30 raw clean images. Six different types of distortions are added to the clean images, resulting in 866 distorted images in total. The quality score of each sample is also provided. 

\textbf{TID2013} is an expansion of Tampere Image Dataset 2008 (TID2008) \cite{ponomarenko2009tid2008}, based on LIVE dataset \cite{sheikh2006statistical}, with the increase of distorted level to 5 and type to 24. Thus, 3000 distorted image samples are generated from 25 original samples. 

\subsubsection{Training Details}

\begin{figure}[t]
    \centering
    \includegraphics[width=1.0\linewidth]{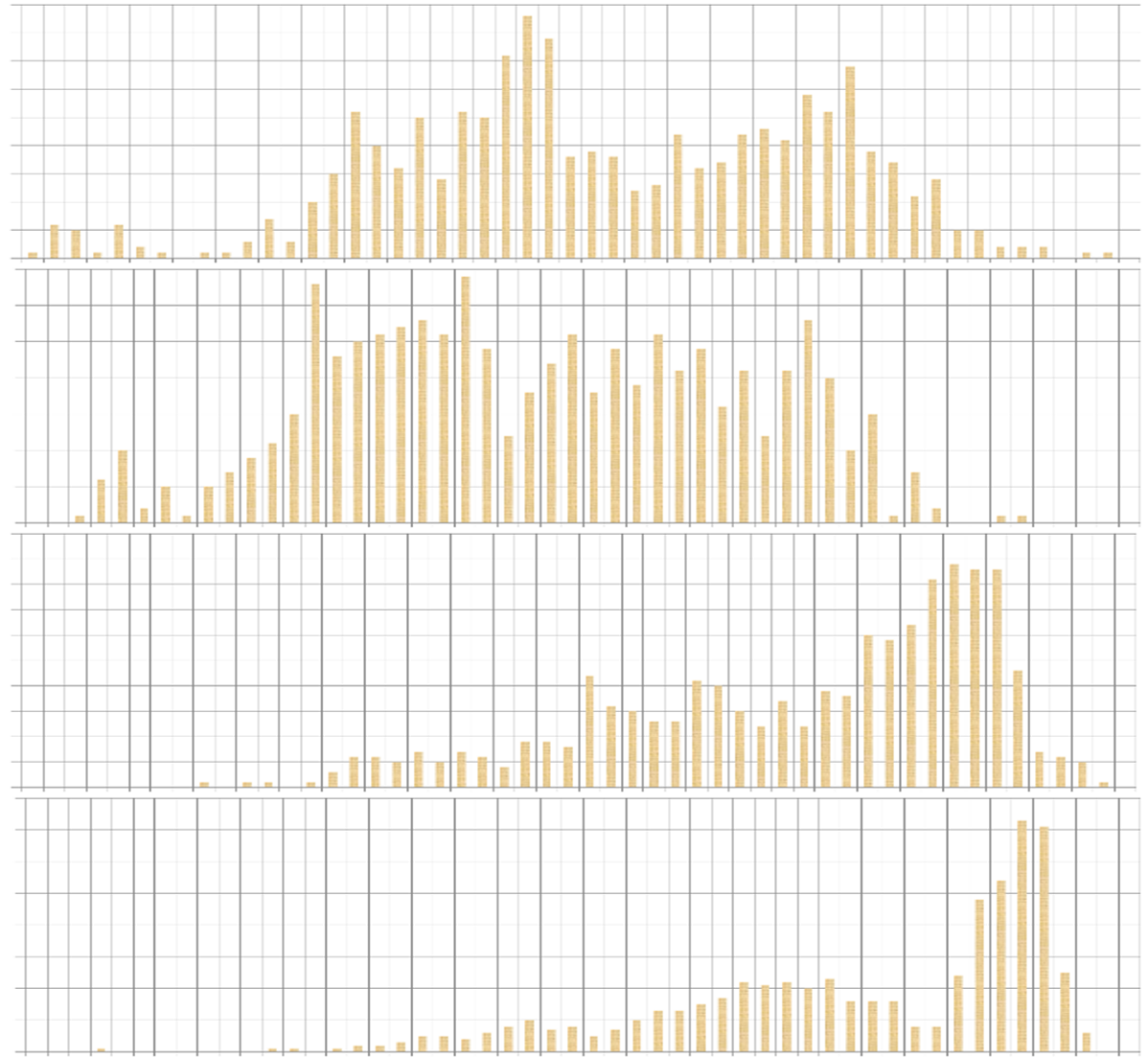}
    \caption{An example of the score distributions from four annotators of the TID2013 dataset.}
    \label{fig:distribution}
\end{figure}

To mimic the annotated scores generated from multiple annotators with different biases, we randomly shift the mean of provided ground truth scores for individual annotators. Meanwhile, we also perturb scores locally with the relative pairwise ranking unchanged. We normalize the scores in the range [-1, 1] for the face expression score prediction task and [0, 1] for the IQA task as the final annotated scores, respectively. We show an example of the augmented score distributions from four annotators of the TID2013 dataset in Fig.~\ref{fig:distribution}. We can see that there are large biases among different annotators.

For the face expression prediction task, we apply ResNet-50 \cite{he2016deep} as our backbone and use the stochastic gradient descent (SGD) optimizer with a momentum of 0.9 and a weight decay of 5$\times 10^{-4}$. For both datasets, scores of valence are used for regression. 
To avoid the influence by the background, for the AFEW-AV dataset, face images are cropped and aligned by the landmarks provided by the dataset, and then resized to 112 $\times$ 112, 
while the detected and aligned face samples in the Aff-Wild dataset are applied directly. During the training phase, we rotate the input images within $\pm 45^{\circ}$ for augmentation. 
We train ResNet-50 on AFEW-AV for 400 epochs with an initial learning rate of $10^{-3}$, and for 800 epochs with the same initial learning rate on Aff-Wild. For both datasets, we use a batch size of 512. 

For the image quality assessment task, samples are randomly cropped and resized to 224 $\times$ 224 with a batch size of 64 for training, while the testing samples are resized to 256 $\times$ 256 and centrally cropped to 224 $\times$ 224. During the training phase, we augment the input samples within $\pm 45^{\circ}$ rotation. The same as that in facial expression prediction tasks, we also apply ResNet-50 as our backbone and use SGD optimizer with a momentum of 0.9 and a weight decay of $5 \times 10^{-4}$. The initial learning rates for the CSIQ dataset and TID2013 are both $10^{-3}$ for 300 epochs. For both of the two tasks, one Nvidia RTX 3090 GPU is used for training. The method is implemented with the Pytorch toolbox \cite{paszke2019pytorch}.

\subsection{Evaluation Criteria}
We adopt three evaluation metrics, namely pairwise ranking accuracy (PRA), Spearman’s rank order correlation coefficient (SROCC), and Pearson’s linear correlation coefficient (PLCC). Since annotation is inconsistent for comparison across different annotators, we evaluate the results on individual annotators and take the average as the final results.

\subsubsection{Pairwise Ranking Accuracy (PRA)}
PRA compares the relative pairwise ranking of the predicted scores with annotated labels. For $N$ samples, we have $N(N-1)/2$ pairs in total. We treat a pair as a correct prediction if the pairwise ranking of the predicted scores from the two samples follows the same relative ranking as labeled annotations. Then the PRA is calculated as the total accuracy given all the possible pairs.

\subsubsection{Spearman’s Rank Order Correlation Coefficient (SROCC)} The SROCC is defined as follows,
\begin{equation}
    \text{SROCC = $1-\frac{6\sum_{i=1}^Nd_i^2}{N(N^2-1)}$},
\end{equation}
where \text{$d_{i}$} denotes the difference between the ranks of $i$-th testing images in the ground truth and predicted quality scores.

\subsubsection{Pearson’s Linear Correlation Coefficient (PLCC)}

PLCC calculates the linear correlation between the annotation $\Tilde{y}$ and the predicted label $y$. The definition is shown as follow,
\begin{equation}
    \text{PLCC = $\frac{\sum_{i=1}^N(\Tilde{y}_i-\mu_{\Tilde{y}})\times({y_i}-\mu_{{y}})}{\sqrt{\sum_{i=1}^N(\Tilde{y}_i-\mu_{\Tilde{y}}^2)}\times\sqrt{\sum_{i=1}^N({y}_i-\mu_{{y}})^2}}$},
\label{eq11}
\end{equation}
where $\mu$ represents the sample mean. The range of PLCC is from -1 to 1, and higher scores mean better performance.

\subsection{Main Results}

\noindent \textbf{Evaluation on Facial Expression Score Prediction.} 
\noindent To validate the effectiveness of our proposed DCR framework, we compare with the regression baseline, \textit{i.e.}, a direct regression with mean squared error (MSE) as the loss function. The results are shown in Table~\ref{tab:facial}, where the top and the bottom part shows the results on the AFEW-AV and Aff-Wild datasets, respectively. $\text{DCR}_{[-]}$ represents the DCR method without GRL in the training. The best results are marked in bold. As expected, \text{DCR} achieves the best performance.


\begin{table}[t]
\centering
\begin{tabular}{L{1.7cm}L{1.3cm}L{1.3cm}L{1.3cm}}
    \toprule
    \text{Method} & PRA (\%) & SROCC & PLCC \\
    \midrule
    \text{Baseline} & 63.38 & 0.369& 0.371\\
    \text{DCR}$_{[-]}$ & 89.10 & 0.868& 0.881 \\
    \text{DCR} & \textbf{89.22} & \textbf{0.875} & \textbf{0.890} \\
    \midrule
    \text{Baseline} & 62.28 & 0.336 & 0.348\\
    \text{DCR}$_{[-]}$ & 83.35 & 0.812 & 0.827\\
    \text{DCR} & \textbf{84.01} & \textbf{0.829} & \textbf{0.837} \\
    \bottomrule
\end{tabular}
\caption{Comparison of different approaches on AFEW-AV (top three) and Aff-Wild (bottom three) datasets. The best performance is marked in bold.}
\label{tab:facial}
\end{table}

\noindent \textbf{Evaluation on Image Quality Assessment.} 
\noindent Two state-of-the-art (SOTA) methods for blind image quality assessment, including MetaIQA \cite{zhu2020metaiqa} and HyperIQA \cite{su2020blindly}, are chosen to compare with our proposed approach. The results on the CSIQ and TID2013 datasets are shown in Table~\ref{tab:CSIQ}. MetaIQA+DCR and HyperIQA+DCR stand for approaches with the same architecture used in the original methods with the combination of the proposed DCR framework. We can see that there is a consistent improvement on all three metrics with DCR adopted in the training.

\begin{table}[t]
\centering
\begin{tabular}{L{2.5cm}L{1.3cm}L{1.3cm}L{1.3cm}}
    \toprule
    \text{Method} & PRA (\%) & SROCC & PLCC \\
    \midrule
    \text{MetaIQA} \cite{zhu2020metaiqa} & 92.46& 0.897& 0.918\\
    \text{MetaIQA}+\text{DCR} & \textbf{93.91} & \textbf{0.916} & \textbf{0.932}\\
    \text{HyperIQA} \cite{su2020blindly} & 90.06& 0.887 & 0.893\\
    \text{HyperIQA}+\text{DCR} & 91.01& 0.891 &0.903\\
    \midrule
    \text{MetaIQA} \cite{zhu2020metaiqa} &83.40 &0.807 & 0.829 \\
    \text{MetaIQA}+\text{DCR} &\textbf{85.59} &\textbf{0.825} & \textbf{0.835}\\
    \text{HyperIQA} \cite{su2020blindly} &81.01 &0.782 & 0.794 \\
    \text{HyperIQA}+\text{DCR} & 81.87&0.812 & 0.828\\
    \bottomrule
\end{tabular}
\caption{Comparison with SOTA methods on the CSIQ (top four) and TID2013 (bottom four) datasets, with the best performance marked in bold.}
\label{tab:CSIQ}
\end{table}


\subsection{Ablation Study}

\noindent \textbf{Study on the Regularization Loss.}
\noindent To evaluate the effectiveness of the regularization term in DCR, we conduct experiments on AFEW-AV and TID2013 with different choices of $\lambda_1$ from 0 to 0.3. The results are shown in Table~\ref{tab:reg_loss}, where the top and the bottom part shows results on AFEW-AV and TID2013, respectively.
From Table~\ref{tab:reg_loss}, we achieve significantly higher performance with the regularization term than the method without regularization, \textit{i.e.}, $\lambda_1=0.0$, on both the two datasets. This indicates that the model is easy to collapse to a naive solution without regularization. The best performance is achieved when $\lambda_1$ is around 0.1$\sim$0.2. 

\begin{table}[t]
\centering
\begin{tabular}{L{1.5cm}L{1.3cm}L{1.3cm}L{1.3cm}}
    \toprule
    $\lambda_1$ & PRA (\%) & SROCC & PLCC \\
    \midrule
    0.0 & 63.15 & 0.311 &0.323 \\
    0.1 & \textbf{89.22} & \textbf{0.875} & \textbf{0.890}\\
    0.2 & 88.51& 0.864& 0.876\\
    0.3 & 86.14& 0.842 & 0.853 \\
    \midrule
    0.0 & 62.47 & 0.337 & 0.349\\
    0.1 & 71.78 & 0.697 & 0.701 \\
    0.2 & \textbf{72.36} & \textbf{0.706} & \textbf{0.712} \\
    0.3 & 67.33& 0.435& 0.446\\
    \bottomrule
\end{tabular}
\caption{Regularization loss on AFEW-AV and TID2013 datasets, respectively, with the best performance marked in bold.}
\label{tab:reg_loss}
\end{table}

\noindent \textbf{Study on the Number of Annotators.}
\noindent Table \ref{tab:annotator} shows the comparison of the AFEW-AV and TID2013 datasets between different numbers of annotators, ranging from one to four. As expected, the best performance is achieved with a single annotator since there is no inconsistent label issue with only one annotator. The performance does not degrade significantly when increasing the number of annotators, showing that it is still tolerant when combining disjoint annotations for training in practical situations. 

\begin{table}[t]
\centering
\begin{tabular}{L{1.5cm}L{1.3cm}L{1.3cm}L{1.3cm}}
    \toprule
    \#Annotator & PRA (\%) & SROCC & PLCC \\
    \midrule
    1 & \textbf{89.31} & \textbf{0.886} & \textbf{0.912}\\
    2 & 88.67&0.876 & 0.897\\
    3 & 88.98 & 0.868& 0.887\\
    4 &89.22 &0.875 & 0.890\\
    \midrule
    1 & \textbf{72.63} & \textbf{0.769} & \textbf{0.789}\\
    2 & 69.06 & 0.705&0.711 \\
    3 & 70.80& 0.749&0.761 \\
    4 & 72.36 & 0.706 & 0.712\\
    \bottomrule
\end{tabular}
\caption{Results with different number of annotators on AFEW-AV and TID2013 datasets, respectively, with the best performance marked in bold.}
\label{tab:annotator}
\end{table}

\noindent \textbf{Study on the Disjoint Ranking Loss.}
\noindent Table \ref{tab:mask} shows the comparison between the margin ranking loss $\mathcal{L}_{\text{rank}}$ and the disjoint margin ranking loss $\mathcal{L}_{\text{d-rank}}$ on the AFEW-AV and TID2013 datasets, respectively. 
Since $\mathcal{L}_{\text{rank}}$ does not consider the inter-annotator biases, it is expected that $\mathcal{L}_{\text{d-rank}}$ outperforms $\mathcal{L}_{\text{rank}}$ with a large margin, \textit{i.e.}, a $19.42\%$ and $6.72\%$ improvement on the PRA for the AFEW-AV and TID2013 datasets, respectively. 

\begin{table}[t]
\centering
\begin{tabular}{L{1.5cm}L{1.3cm}L{1.3cm}L{1.3cm}}
    \toprule
    Loss & PRA (\%) & SROCC & PLCC \\
    \midrule
    $\mathcal{L}_{\text{rank}}$ &69.80 & 0.486& 0.513 \\
    $\mathcal{L}_{\text{d-rank}}$ & \textbf{89.22} & \textbf{0.875} & \textbf{0.890} \\
    \midrule
    $\mathcal{L}_{\text{rank}}$ &65.64 &0.412 & 0.423\\
    $\mathcal{L}_{\text{d-rank}}$ & \textbf{72.36} & \textbf{0.706} & \textbf{0.712} \\
    \bottomrule
\end{tabular}
\caption{Comparison between margin ranking loss $\mathcal{L}_{\text{rank}}$ and disjoint margin ranking loss $\mathcal{L}_{\text{d-rank}}$ on AFEW-AV and TID2013 datasets, respectively, with the best performance marked in bold.}
\label{tab:mask}
\end{table}

\noindent \textbf{Study on Variant Margins.}
\noindent We also test the role of margin in our proposed disjoint ranking loss. We vary the margin \text{$\gamma$} from 0 to 0.2 on the AFEW-AV and TID2013 datasets, as shown in Table \ref{tab:margin}. For both of the two datasets, we achieve the best results when $\gamma=0$, \textit{i.e.}, no margin is used. With the increase of the margin, the performance keeps degrading. Generally, the margin is more appropriate in the classification task where samples from different classes need to be separated as much as possible. However, for regression tasks, two similar inputs should generate two similar scores, rather than intentionally enlarging the distance of the predicted scores. As a result, setting $\gamma=0$ is more suitable for regression tasks.

\begin{table}[!t]
\centering
\begin{tabular}{L{1.5cm}L{1.3cm}L{1.3cm}L{1.3cm}}
    \toprule
    Margin $\gamma$ & PRA (\%) & SROCC & PLCC \\
    \midrule
    0.0 & \textbf{89.22} & \textbf{0.875} & \textbf{0.890} \\
    0.1 &88.43 & 0.841 & 0.887 \\
    0.2 &87.35 &0.857 & 0.885 \\
    \midrule
    0.0 & \textbf{72.36} & \textbf{0.706} & 0.712\\
    0.1 & 71.40&0.702 & \textbf{0.716}\\
    0.2 & 69.64 & 0.679& 0.688\\
    \bottomrule
\end{tabular}
\caption{Results with variant margins in our disjoint ranking loss on the AFEW-AV and TID2013 datasets, respectively, with the best performance marked in bold.}
\label{tab:margin}
\end{table}

\section{Conclusion}

In this paper, we propose a novel disjoint contrastive regression (DCR) framework
to address the challenges of the inconsistent label problems due to the annotation bias. 
Experimental results with extensive studies demonstrate the effectiveness of our proposed DCR framework. In our future work, we will consider the video-based regression tasks with weak labels for disjoint annotations.

\bibliographystyle{IEEEbib}
\bibliography{icme2022template}

\end{document}